\documentclass[pmlr,twocolumn,10pt]{jmlr} 

\usepackage{booktabs}
\usepackage{siunitx}

\usepackage{algorithm2e}
\usepackage{float}
\usepackage{svg}
\usepackage{multirow}
\usepackage{url}
\usepackage{colortbl}
\usepackage{soul}          
\usepackage{color}
\usepackage{hyperref}
\usepackage{adjustbox}
\def\mybar[#1]#2{
  {\color{black}\rule[0.1ex]{#1mm}{5pt}} #2}
\graphicspath{{figures/}}

\theorembodyfont{\upshape}
\theoremheaderfont{\scshape}
\theorempostheader{:}
\theoremsep{\newline}

\jmlrvolume{LEAVE UNSET}
\jmlryear{2023}
\jmlrsubmitted{LEAVE UNSET}
\jmlrpublished{LEAVE UNSET}
\jmlrworkshop{Conference on Health, Inference, and Learning (CHIL) 2023} 

\title[Life event detection]{Rare Life Event Detection via Mobile Sensing Using Multi-Task Learning}

\author{%
\Name{Arvind Pillai} \Email{arvind.pillai.gr@dartmouth.edu}\\
\Name{Subigya Nepal} \Email{sknepal@cs.dartmouth.edu}\\
\Name{Andrew Campbell} \Email{andrew.t.p.campbell@gmail.com}\\
\addr Dartmouth College, USA
}

\begin{document}

\maketitle

\begin{abstract}
Rare life events significantly impact mental health, and their detection in behavioral studies is a crucial step towards health-based interventions. We envision that mobile sensing data can be used to detect these anomalies. However, the human-centered nature of the problem, combined with the infrequency and uniqueness of these events makes it challenging for unsupervised machine learning methods. In this paper, we first investigate granger-causality between life events and human behavior using sensing data. Next, we propose a multi-task framework with an unsupervised autoencoder to capture irregular behavior, and an auxiliary sequence predictor that identifies transitions in workplace performance to contextualize events. We perform experiments using data from a mobile sensing study comprising N=126 information workers from multiple industries, spanning 10106 days with 198 rare events ($<2\%$). Through personalized inference, we detect the exact day of a rare event with an F1 of 0.34, demonstrating that our method outperforms several baselines. Finally, we discuss the implications of our work from the context of real-world deployment.

\end{abstract}
\paragraph*{Data and Code Availability.}
Tesserae study data \citep{mattingly2019tesserae} can be obtained through a data usage agreement (\url{https://tesserae.nd.edu/}). Code is not publicly available, but provided on request. 


\paragraph*{Institutional Review Board (IRB).}
The study protocol is fully approved by the Institutional Review Boards at Dartmouth College, University of Notre Dame, University of California-Irvine, Georgia Tech, Carnegie Mellon University, University of Colorado-Boulder, University of Washington, University of Texas-Austin, and Ohio State University. This research is based upon work supported in part by the Office of the Director of National Intelligence (ODNI), Intelligence Advanced Research Projects Activity (IARPA), via IARPA Contract No. 2017-17042800007. The views and conclusions contained herein are those of the authors and should not be interpreted as necessarily representing the official policies, either expressed or implied, of ODNI, IARPA, or the U.S. Government.
\section{Introduction}
Life events (LE) are significant changes in an individual's circumstances that affect interpersonal relationships, work, and leisure \citep{hill2002mcgraw}. The nature of these events inevitably affect mental well-being and general health \citep{goodyer2001life}. The detrimental effects of adverse LEs (e.g., death of a loved one, losing a job, terminal illness) have been widely studied and known to be associated with a higher incidence of depression and altered brain network structure \citep{falkingham2020accumulated, gupta2017early}. In contrast, positive LEs (e.g., taking a vacation, job promotion, childbirth) are associated with increased life satisfaction, and higher cognitive function \citep{castanho2021association}. Moreover, LEs affect cardiovascular vascular disease risk factors, such as increased central adiposity, heightened exposure to inflammation, and elevated resting blood pressure \citep{steptoe2013stress}. A study by \citet{steptoe2012stress} suggests that even minor LEs can trigger cardiac events like myocardial ischemia. 

 Sensing data is a multivariate time series, and traditional ML approaches for anomaly detection include the One-Class SVM (OCSVM) \citep{ma2003time}, Isolation Forest (IF) \citep{liu2008isolation}, and logistic regression \citep{hilbe2009logistic}. However, these approaches do not capture temporal dependencies. Additionally, creating user-specific thresholds is critical in human-centered tasks. Thus, methods which directly predict anomalies (IF) or require threshold parameter tuning (OCSVM) are not ideal. 

Recently, timeseries based autoencoders have received significant attention \citep{rumelhart1985learning, zhou2017anomaly, audibert2020usad,  su2019omni}, and many methods use RNN variants such as LSTM \citep{hochreiter1997long} and GRU \citep{cho2014properties} to capture temporal dependencies. In an autoencoder, the reconstruction error is used to detect anomalies. However, the complexity of human behavior and class imbalance creates biased learned representations, making it challenging to distinguish normal and rare events \citep{pang2021deep}. An intuitive solution to this problem involves increasing the error of rare events without affecting normal events. Multi-task learning achieves this by incorporating additional information from a related task(s) to learn a shared representation, and thus compensate for the limitations of a purely unsupervised approach \citep{vidya2019canbus, wu2021multi}.

In this paper, we first use statistical analysis to examine whether LEs result in behavioral shifts observable through mobile sensing. Next, we propose Multi-Task Anomaly Detection (MTAD) to detect ``in-the-wild" rare LEs using behavioral mobile sensing data. We hypothesize that a standalone unsupervised autoencoder cannot effectively capture differences between normal and rare events using the reconstruction error because of the heterogeneity in human data and the significant class imbalance ($<2\%$ rare events). Thus, MTAD trains an auxiliary sequence predictor to contextualize events by capturing changes in workplace performance due to the event. For example, a participant reported that getting promoted positively impacted their work performance, while another person mentioned that visiting their sick parents had a negative effect. We aim to identify such changes at inference and compute a scaling factor that magnifies the reconstruction error of rare events.

\subsection{Contributions}
Toward the vision of detecting life events from mobile sensing data collected in in-the-wild settings, our main contributions are as follows. First, we perform granger-causality testing to detect if the days before an event can be used to predict days after the event. Thus, establishing a relationship between LEs and behavior (section \ref{sec:statisticaltesting}). Second, we propose a multi-task learning architecture consisting of two components: (1) an LSTM based encoder-decoder to calculate an anomaly score, and (2) a sequence predictor to contextualize the anomaly score by inferring a transition in workplace performance (section \ref{sec:method}). Third, we perform empirical analysis on a real-world dataset to compare MTAD with five state-of-the-art baselines to analyze performance and robustness (section \ref{sec:results}). Finally, we rigorously evaluate parameters that affect MTAD (section \ref{sec:results}) and discuss implications of our research (section \ref{sec:disc}).
    
\section{Related Work} \label{sec:related}

\subsection{Change Point Detection}
Change point detection (CPD) refers to the change in the state of a time series. CPD has been applied to various physiological signals \citep{shvetsov2020unsupervised, fotoohinasab2020graph, stival2022doubly}. For example, \citet{shvetsov2020unsupervised} propose an unsupervised approach based on point clouds and Wasserstein distances to detect six different types of arrhythmias. Designing appropriate metrics that can identify change is vital in CPD. Consequently, \citet{chen2019automated} design a metric for EEG CPD by modifying several similarity metrics from other domains. In fitness, \citet{stival2022doubly} propose a method combining CPD and Gaussian state space models to detect actionable information such as physical discomfort and de-training.  

\subsection{Anomaly detection methods} Approaches for anomaly detection in multivariate time series are varying, and several challenges exist based on the method, and applied area \citep{pang2021deep}. In deep learning, the LSTM encoder-decoder (or LSTM autoencoder) has received a lot of attention. \citet{malhotra2016lstm} demonstrate the robustness of using an LSTM in an autoencoder framework. Similarly, \citet{park2018multimodal} propose the LSTM-VAE, which combines the temporal modeling strength of LSTM with the variational inference capability of a VAE. The resulting model obtains better generalization for multimodal sensory signals. The Deep Autoencoding Gaussian Mixture Model (DAGMM) jointly optimizes two components to enable optimal anomaly detection, an autoencoder computes a low-dimensional representation, and a gaussian mixture model that predicts sample membership from the compressed data \citep{zong2018deep}. \citet{audibert2020usad} propose an adversarially trained autoencoder framework to detect anomalies. For anomalous driving detection, \citet{vidya2019canbus} introduce a multi-task architecture that magnifies rare maneuvers using domain knowledge regarding the frequency of driving actions.

\subsection{Life event detection} To the best of our knowledge, there are two works similar to ours. First, \citet{faust2021examining} use OCSVM to assess the response to an adverse life event using physiological and behavioral signals from wrist-worn wearables. They focus on examining the responses to the adverse event and the coping strategies employed by the participant. Their findings suggest the existence of behavioral deviations after a negative event, motivating us to focus on prediction. Second, \citet{burghardt2021having} detect abnormal LEs using wrist-worn wearables from hospital and aerospace workers. Their method works by first creating a time series embedding using a hidden markov model variant and then uses a random forest or logistic regression for classification.

Our work differs from the previous studies in several ways: (1) we use smartphone behavioral data instead of wearable physiological data, (2) we consider postive, negative, and multiple LEs (differing from \citet{faust2021examining}), (3) we focus on deep models instead of traditional ML, (4) our data has an extremely low anomaly ratio ($<2\%$) compared to \citet{burghardt2021having} (11.7\% and 14.9\%). Thus, we view our problem to be significantly challenging. Moreover, we provide crucial statistical motivation to pursue LE detection.


\section{Study} \label{sec:study}
The Tesserae study \citep{mattingly2019tesserae} recruited 757 information workers across different companies in the USA for one year where participants respond to several surveys. They were instrumented with a Garmin vivoSmart 3 wearable and a continuous sensing app is installed on their phone. Participants are instructed to maintain data compliance level of 80\% to warrant eligibility for monetary remuneration. The sub-cohort's age ranged from 21 to 64, with an average of 34. Of the 126 individuals, the dataset is fairly balanced with 67 and 59 identified as male and female, respectively. The top 3 areas of occupation were Computer and Mathematical Sciences, Business and Finance, and Engineering. Roughly 98\% of the participants had at least a college degree. In terms of mobile platform, the cohort had 66 Android users and 60 iOS users. Please refer to the link in the data availability statement to learn about the Tesserae study. Additional demographic information is listed in Appendix \ref{apd:second}.

\subsection{Features}
In contrast to studies using wearable physiology data \citep{faust2021examining, burghardt2021having}, we use daily summaries of behavioral mobile sensing features in our analyses. Overall, we used walking duration,  sedentary duration, running duration, distance traveled, phone unlock duration, number of phone unlocks, number of locations visited, and number of unique locations visited. Further, to better understand user behavior, we divide the features (except number of unique locations visited) into 4 ``epochs" for modelling: epoch 1 (12am - 6am), epoch 2 (6am - 12pm), epoch 3 (12pm - 6pm), and epoch 4 (6pm - 12am). Ultimately, 29 features were used for analyses. Previous studies elucidate the importance of these features to understand human behavior from a workplace context \citep{nepal2020detecting, mirjafari2019differentiating}.

\subsection{Ground Truth} \label{sec:ground}


The definition of a \textbf{significant life event} is subjective and depends on the individual. We adopt a widely accepted definition from job stress literature, which describes these events as situations where psychological demand exceeds available resources \citep{karasek1981job}.

After study completion, participants were asked to describe significant LEs using their diaries, calendars, and other documents. Participants provided free-text descriptions for every event, start and end dates, date confidence, significance, valence (positive/negative), type of event, and workplace performance impact. Valence, date confidence, and workplace performance are reported on a 1-7 likert scale as follows: (1) Valence: ``1" indicated ``Extremely Positive" and ``7" indicated ``Extremely Negative", and (2) Date confidence: ``1" indicated ``Lowest confidence" and ``7" indicated ``Highest confidence". Workplace performance impact is assigned to one of the following - ``Large Negative Effect", ``Medium Negative Effect", ``Small Negative Effect", ``No Effect", ``Small Positive Effect", ``Medium Positive Effect", ``Large Positive Effect".

 Our selection criteria is as follows: (1) Valence must be Extremely Positive, Very Positive, Very Negative, or Extremely Negative, and (2) Date confidence must be ``Moderately High", ``High", or ``Highest". Next, we set a 30-day date limit before and after an event for analysis. For overlapping events, the limit is set 30 days before and after the first and last events, respectively. These choices are based on a study by \citet{faust2021examining} examining the impact of LEs within a 60-day period. Finally, the missingness and uneven spacing (discontinuous days) within this time frame must be $<25\%$. Every day was labelled as ``1" indicating a rare event or ``0" indicating a normal event. For workplace performance, the label is forward filled, and an ``Unknown" label is assigned to days before the rare event. Our final dataset consists of 10106 days from 126 participants with 198 rare LEs ($<2\%$). 

\section{Statistical Testing} \label{sec:statisticaltesting}
\begin{figure}[h]
  \centering
  \includegraphics[width=230pt]{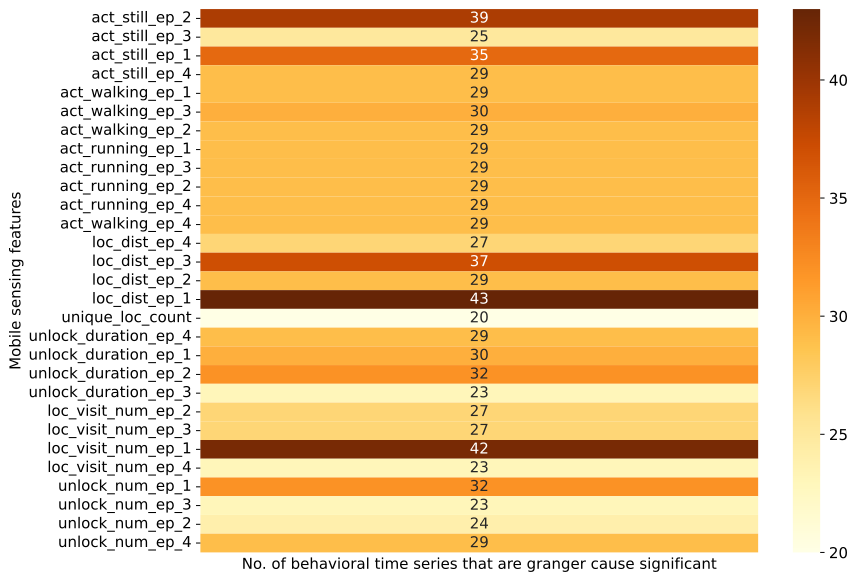}
  \caption{Heatmap indicating the number of time series (out of 159) that are granger-cause significant at $p<0.05$. Larger values imply that the corresponding feature changes after an LE in many participants, and could be important for detection. The x-axis is the number of significant time series, i.e., the count of significant time series out of 159. The y-axis are the mobile sensing features. ``act" is an activity which can be still, walking, running. ``loc" specifies location. ``dist" is distance. ``num" is number.}
    \label{fig:grangercause}
\end{figure}

Initially, we ask the question: ``Does the behavior of an individual change after an LE?". If so, what features are significant in most of the time series? To this end, we applied the granger-causality test as follows. First, we split the 159 multivariate time series (29 features) into two parts, one before and including the rare event ($T^{pre}$) and the other after the rare event ($T^{post}$). Next, for each feature, a granger-cause test is applied to investigate if $T^{pre}$ granger-causes $T^{post}$. A $p<0.05$ implies the time series for the specific feature is significant. Finally, we sum the significant time series across each feature. For example, in Figure \ref{fig:grangercause}, loc\_visit\_num\_ep\_1 has a value of 42, which implies that 42 out of 159 time series were granger-cause significant. We used the statsmodel package for python to apply the granger-cause test with a lag of upto 6. The SSR-based F-test is used to evaluate statistical significance at $p=0.05$. Significance at any lag is considered granger-causality, and the total number of significant time series (out of 159) for each feature is displayed in Figure \ref{fig:grangercause}.

 
 From Figure \ref{fig:grangercause}, we observe that the number of locations visited and location distance between 12am-6am and 12pm-6pm is significantly impacted by LEs in several cases, suggesting location behaviors are crucial to LE detection. In addition, we observe that walking and running have approximately the same number of granger-cause time series across episodes, suggesting an overall change throughout the day rather than a shift in the timing of these activities. In contrast, sedentary action varies across episodes, suggesting that LEs might affect the sedentary state at different times of the day. Also, unlock duration and count vary due to a life event. While the number of unlocks between 12am-6am has the most number of significant time series, the unlock duration between 6am-12pm has more significance comparatively.

\section{Multi-task anomaly detection} \label{sec:method}
\subsection{Problem formulation}\label{sec:problem}
Given a set of $I$ participants, we define a multivariate time series for each user $u \in \{1,\ldots, I\}$ with $T$ days as $\mathcal{T}^u = \{{\boldsymbol{x}^u_{1}},\ldots,\boldsymbol{x}^u_{T}\}$, where $\boldsymbol{x} \in \mathbb{R}^m$ , $m$ is the number of mobile sensing features, and $t \in \{1,\ldots,T\}$ is a specific day.
To model temporal dependencies, we apply a rolling window approach to the time series. A window at day $t$ with a predefined length $l$ is given by:
\begin{equation}\label{eq:1}
    W_{t} = \{\boldsymbol{x}_{t-l+1}, \ldots, \boldsymbol{x}_{t-1}, \boldsymbol{x}_{t}\}
\end{equation}
Using equation \eqref{eq:1}, the user's multivariate time series $\mathcal{T}^{u}$ is transformed into a collection of windows $\mathcal{W} = \{W_{1}, \dots, W_{T}\}$, where $W \in \mathbb{R}^{l \times m}$. Next, a window is assigned a binary label $y^R\in\{0,1\}$, where $y^R=1$ indicates a rare life event at the time $t$ (i.e., $y_t^R = 1$) and $y^R=0$ indicates a normal event in all other cases. Observe that we only consider the exact day of the rare event to be a rare window. Given that each participant's windows is transformed separately, we generalize the entire collection of normal and rare event windows across participants as $\mathcal{W}_{normal} = \{W_1, \ldots, W_N\}$ and $\mathcal{W}_{rare} = \{W_1, \dots, W_R\}$, respectively.

\begin{figure*}[htbp]
  \centering
  \includegraphics[width=0.7\linewidth]{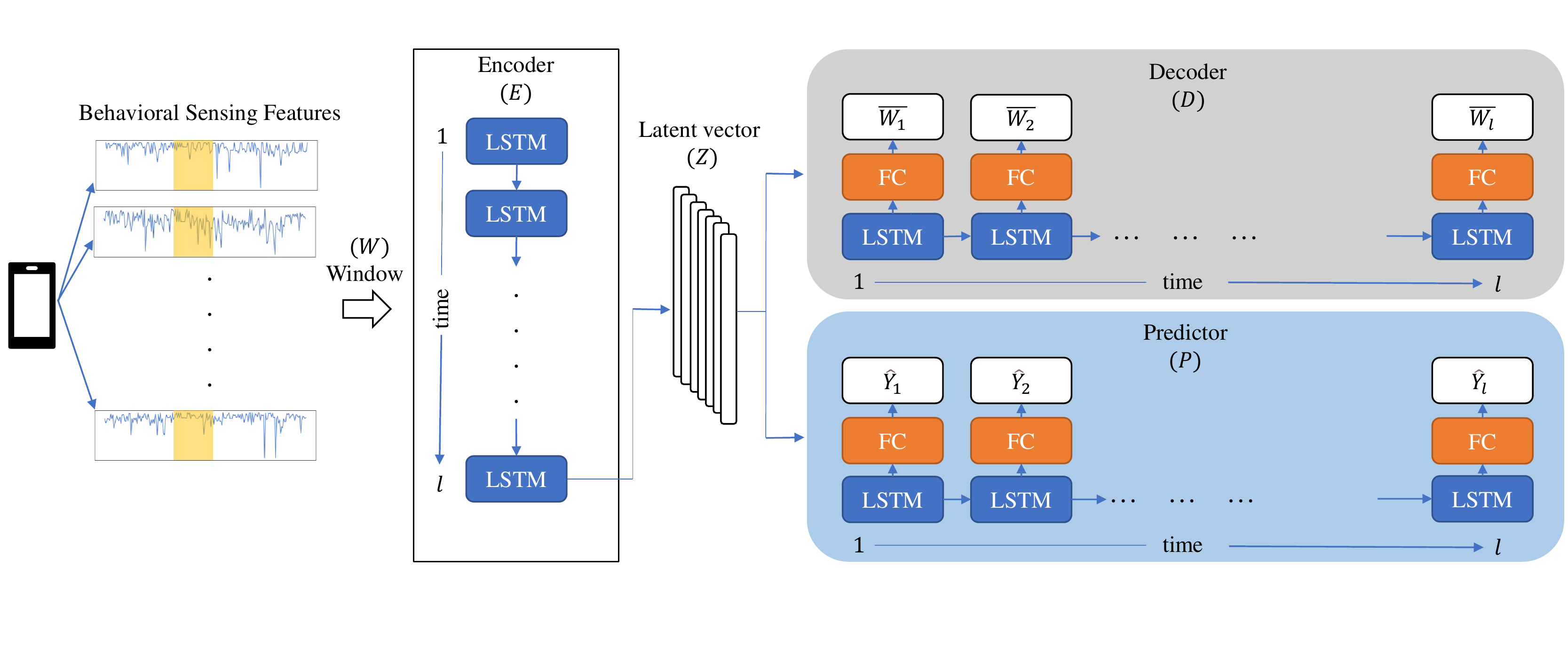}
  \caption{The proposed multi-task learning architecture illustrating training information flow for a window $W$ of length $l$.}
  \label{fig:architecture}
\end{figure*}

In our context, we define a multi-task model with two related tasks trained using $\mathcal{W}_{normal}$. First, an unsupervised learning task trained to reconstruct the input $\mathcal{W}_{normal}$, which  produces higher errors or anomaly scores when reconstructing $\mathcal{W}_{rare}$. Thus, facilitating rare life event detection. Second, a supervised learning task to contextualize and scale the anomaly score. Here, $\mathcal{W}_{normal}$ is trained to predict a workplace performance vector $\boldsymbol{y} \in \mathbb{R}^l$, where each day $t \in \{1, \ldots, l\}$ in $W$ represented by $y_t$ belongs to one of the workplace performance labels described in \ref{sec:ground}.

\textit{Problem Statement.} Given a participant's multivariate time series window $W_t$ and the corresponding workplace performance vector $\boldsymbol{y}$, the objective of our problem is to train a multi-task framework capable of detecting a rare life event at time $t$.
\subsection{Multi-task Architecture}

Our multi-task framework (Figure \ref{fig:architecture}) consists of three components: an encoder E which maps a window $W$ to a low-dimensional representation (latent space) $Z$, a decoder $D$ to reconstruct $W$ from $Z$ (\ref{taska}), and a sequence predictor $P$ to predict the workplace performance vector $\boldsymbol{y}$  (\ref{taskb}).

\subsubsection{Unsupervised Autoencoder (Task A)} \label{taska}

We capture temporal dependencies from the multivariate time series using LSTMs \citep{hochreiter1997long} to build the encoder-decoder architecture. An LSTM encoder learns from an input window $W$ by running through the day-wise input sequence and computes a fixed size latent space $Z$. Next, $Z$ is copied multiple times to match the length of the window. Finally, the LSTM decoder $D$ uses $Z$ to reconstruct the input sequence, and the reconstructed sequence is represented as $\overline{W}$. We train the LSTM encoder-decoder (LSTM-ED) by minimizing the reconstruction error between $W$ and $\overline{W}$ using the mean squared error defined as:
\begin{equation} \label{eq:2}
    \mathcal{L}_A = \frac{1}{l \times m}\|W - \overline{W}\|^2_F
\end{equation}
where $\overline{W}=D(Z)$; $ Z=E(W)$; $\|\cdot\|_F$ is the Frobenius norm

Recall that we only use $\mathcal{W}_{normal}$ to train the LSTM-ED to learn normal event representations. Therefore, by using the reconstruction error as an anomaly score $\alpha$, we can detect rare events based on their higher $\alpha$ values. However, it is possible that some participants or events do not exhibit significant behavior changes which can be captured by our LSTM-ED through $\alpha$. To address this challenge, we attempt to identify anomalies through a supervised learning setup in the next section. \citet{srivastava2015unsupervised} describes LSTM encoder-decoder architectures in detail.

\begin{algorithm2e}[htbp]\label{alg:1}
\caption{Training}
\KwIn{$\mathcal{D}_{train}$ with $\mathcal{W}_{normal} = \{W_1, \ldots, W_N\}$, $\{Y_1, \ldots, Y_N\}$, class weight vector $\boldsymbol{w}$, and number of epochs $E$.}
\KwOut{Trained $E$, $D$, $P$}
$E$, $D$, $P$ $\leftarrow$ initialize weights\;
$e \leftarrow 1$\;
\Repeat{$e=E$}{
\For{$n \leftarrow 1$ to $N$}{
    $Z_n \leftarrow E(W_n)$\;
    $\overline{W_n} \leftarrow D(Z_n)$\;
    $\widehat{Y_{n}} \leftarrow P(Z_n)$\;
    $\mathcal{L}_A \leftarrow \frac{1}{l \times m}\|W_n - \overline{W_n}\|^2_F$\;
    $\mathcal{L}_B \leftarrow - \sum_{i=1}^{l}\sum_{j=1}^{c} Y_{nij} \times \ln(\widehat{Y_{nij}}) \times w_j$\;
    $\mathcal{L} \leftarrow \mathcal{L}_A + \mathcal{L}_B$\;
    $E, D, P \leftarrow$ update weights using $\mathcal{L}$\;
}
$e \leftarrow e + 1$\;
}
\end{algorithm2e}

\subsubsection{Sequence Prediction (Task B)} \label{taskb}
To scale the anomaly score $\alpha$, we train a supervised sequence predictor $P$ to detect day-wise workplace performance. The window $W$ and a true workplace performance label vector $\boldsymbol{y}\in \mathbb{R}^l$ as training inputs, where the label for day $t \in \{1, \ldots, l\}$ in $W$represented by $y_{t}$ has one of the performance labels described in section \ref{sec:ground}. Moreover, $Y \in \mathbb{R}^{l \times c}$ represents one-hot vectors from $\boldsymbol{y}$ with $c$ classes ($c$ = 8). Observe that $W$ is same for tasks A and B. Hence, our architecture shares the LSTM encoder network $E$ described in section \ref{taska}. For Task B, $P$ is composed of an LSTM network to further extract temporal features from the latent representation $Z$ followed by a fully connected layer $FC$ (softmax activation) to predict day-wise class probabilities ${\widehat{Y}}$. At inference, ${\widehat{Y}}$ is mapped to the predicted workplace performance label vector $\boldsymbol{\widehat{y}}$. The model is optimized using the weighted categorical cross-entropy loss function defined by:

\begin{equation} \label{eq:3}
    \mathcal{L}_B = - \sum_{i=1}^{l}\sum_{j=1}^{c} Y_{ij} \times \ln(\widehat{Y_{ij}}) \times {w_j}
\end{equation}
where $\widehat{Y} = \text{softmax}(FC)$; $w_j \text{ is the weight for class } j$.

From equations \eqref{eq:2} and \eqref{eq:3}, we can represent the final loss function as $\mathcal{L} = \mathcal{L}_A + \mathcal{L}_B$. The proposed multi-task architecture is trained in an end-to-end fashion, where the tasks A and B are jointly optimized by minimizing $\mathcal{L}$ (Algorithm \ref{alg:1}).

\subsubsection{Inference}
The detection phase workflow (Algorithm \ref{alg:2}) computes the anomaly score $\alpha$ from Task A and a scaling factor $s$ from Task B for each test window. \\
\noindent\textbf{Anomaly Score.} Recall that, our goal is to identify a rare event on the exact day. Thus, we unroll $W$ and $\overline{W}$ to compute the score for the most recent day $t$ as follows:
\begin{equation} \label{eq:5}
    \alpha = \frac{1}{m}\sum_{j=1}^{m}(x_j - \overline{x_j})^2
\end{equation}
where $\boldsymbol{x} \text{ and } \boldsymbol{\overline{x}}$ are the true and reconstructed multivariate time series, respectively; $m$ is the number of mobile sensing features. \\
\noindent\textbf{Scaling factor.} The effect of stressful life events affect work-life balance in US workers and reduce productivity \citep{hobson2001compelling}. Thus, there is reason to believe workplace performance shifts after an LE. To capture this change, we first transform the predicted workplace performance vector $\boldsymbol{\widehat{y}}$ into a binary vector $\boldsymbol{r} \in \mathbb{R}^{l-1}$ defined as:
\[ r_{t-1} = \begin{cases} 
      1 & \widehat{y}_t \neq \widehat{y}_{t-1} \\
      0 & otherwise \\
   \end{cases}
\]
where $t \in \{2,\ldots,l\}$

In essence, we identify the day of workplace performance change and use it as a proxy for rare event detection. For example, consider a $\boldsymbol{\widehat{y}} =$ [``Unknown", ``Unknown", ``Large Negative Effect"] with $l = 3$, then the transition vector is $r = [0, 1]$. Initially, we assumed that a value ``1" on the most recent day could be directly used to detect the rare event. However, this idea had two major limitations. First, it is possible that larger window sizes might have multiple transitions ($r_t = 1$). Second, erroneous predictions might hinder detection. Thus, we exponential weight our transition vector $r$. Intuitively, more recent workplace performance shifts have larger impact on behavioral changes owing to LE. The scaling factor $s$ aims to capture the abovementioned effect as follows:

\begin{equation}\label{eq:6}
    s = \frac{1}{l-1}\sum_{t=1}^{l-1} e^{-\lambda t r_t}
\end{equation}
where $\lambda$ is a constant decay factor.\\

\noindent\textbf{Detection.} The final scaled anomaly score $\delta$ is computed from equations \eqref{eq:5} and \eqref{eq:6} in the following way:
$\delta = \frac{\alpha}{s} \nonumber$. Observe that $\delta = \alpha$ when $\boldsymbol{r}$ is a zero vector, i.e., a vector with no workplace performance changes. Ultimately, a window $W_t$ with a scaled anomaly score $\delta$ has a rare life event at $t$ ($y^R_t = 1$) if $\delta$ is greater than a threshold $\gamma$. However, the scarcity of rare events hindered threshold tuning based on performance metrics. Thus, $\gamma$ is the 95\textsuperscript{th} percentile anomaly scores from the validation data set.

\begin{algorithm2e}[h]\label{alg:2}
\caption{Inference}
\KwIn{$\mathcal{D}_{test}$ with $\mathcal{W} = \{W_1, \ldots, W_{N+R}\}$, $\gamma$ from $D_{val}$, $l$, $\lambda$.}
\KwOut{$\boldsymbol{y^R} = \{y^R_1, \ldots, y^R_{N+R}\}$}
\For{$n \leftarrow 0$ to $N + R$}{
    $Z_n \leftarrow E(W_n)$\;
    $\overline{W_n} \leftarrow D(Z_n)$\;
    $\widehat{Y_{n}} \leftarrow P(Z_n)$\;
    $\boldsymbol{x}, \overline{\boldsymbol{x}} \leftarrow$ unroll($W_n$), unroll($\overline{W_n}$) \;
    $\alpha \leftarrow \frac{1}{m}\sum_{j=1}^{m}(x_j - \overline{x_j})^2$\;
    $\widehat{\boldsymbol{y}} \leftarrow$ get class labels from probabilities $\widehat{Y_n}$\;
    $\boldsymbol{r} \leftarrow$ compute binary transition vector from $\widehat{\boldsymbol{y}}$\;
    $s \leftarrow \frac{1}{l-1}\sum_{t=1}^{l-1} e^{- \lambda t r_t}$\;
    $\delta \leftarrow \frac{\alpha}{s}$\;
    \eIf{$\delta > \gamma$}{
        $y^R_n \leftarrow 1$\;
    }{$y^R_n \leftarrow 0$\;}
    }
\end{algorithm2e}

\subsection{Experimental Setup} \label{sec:experiment}
We performed several pre-processing steps for optimal training and inference. First, we impute missing data and unevenly sampled time series using the mean. Second, we forward fill missing rare event and workplace performance labels. Third, within-subject feature normalization is applied as a precursor for personalization. Finally, each participant's time series is transformed into windows of length $l=10$ using equation \eqref{eq:1}.

In our analysis, we divide $\mathcal{W}_{normal}$ and the corresponding $y$ into training ($\mathcal{D}_{train}$), validation ($\mathcal{D}_{val}$), and test ($\mathcal{D}_{test}$) with ratio of 80:10:10, respectively. Next, we append $\mathcal{W}_{rare}$ to $\mathcal{D}_{test}$. Note that rare events are held-out for testing and are neither used for training nor validation. Moreover, we generate ten different user-dependent splits to ensure the training, validation, and test data sets consist of time series from all participants. To assess positive class performance in imbalanced data, we use precision (P), recall (R), and the F1-score, and report mean and standard deviation across the splits. The windows are unrolled at inference to detect a rare event on the exact day. Thus, identifying if a rare event is present within a window is not considered to be an accurate detection. 

\subsection{Results}\label{sec:results}
In this section, we examine the properties of the proposed framework by comparing it with other baselines (\ref{performance}), analyzing the strengths of personalization (\ref{personalresults}), and estimating the changes in performance at different window sizes and decay constants for the scaling factor (\ref{effectofparameters}). Next, we perform an ablation study to assess the necessity of the sequence predictor (\ref{ablation}). Finally, we assess the types of events identified by MTAD (\ref{eventtype}).

\begin{figure*}[h]
  \centering
  \includegraphics[width=0.8\linewidth]{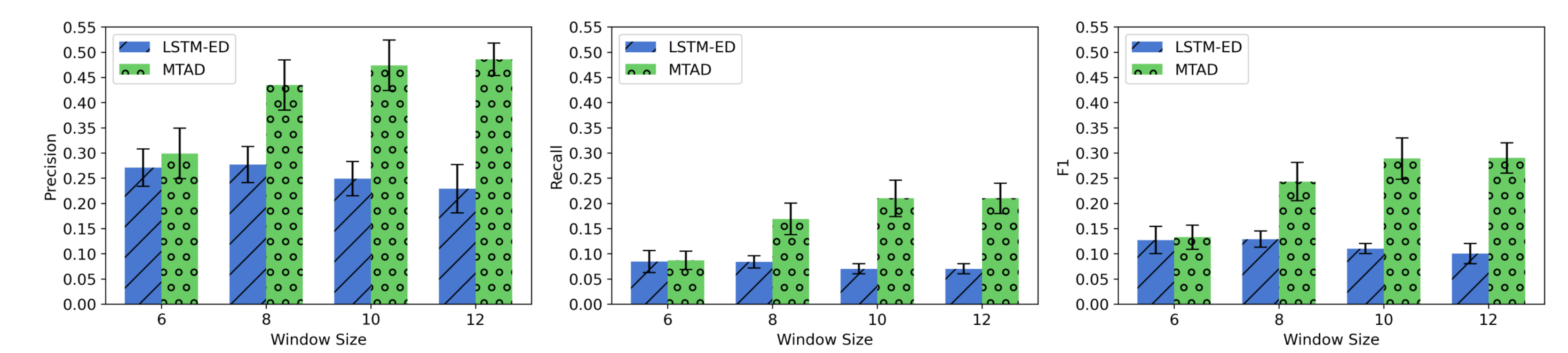}
  \caption{Comparison between general LSTM-ED and MTAD at different window sizes.}
  \label{general}
\end{figure*}
\begin{figure*}[h]
  \centering
  \includegraphics[width=0.8\linewidth]{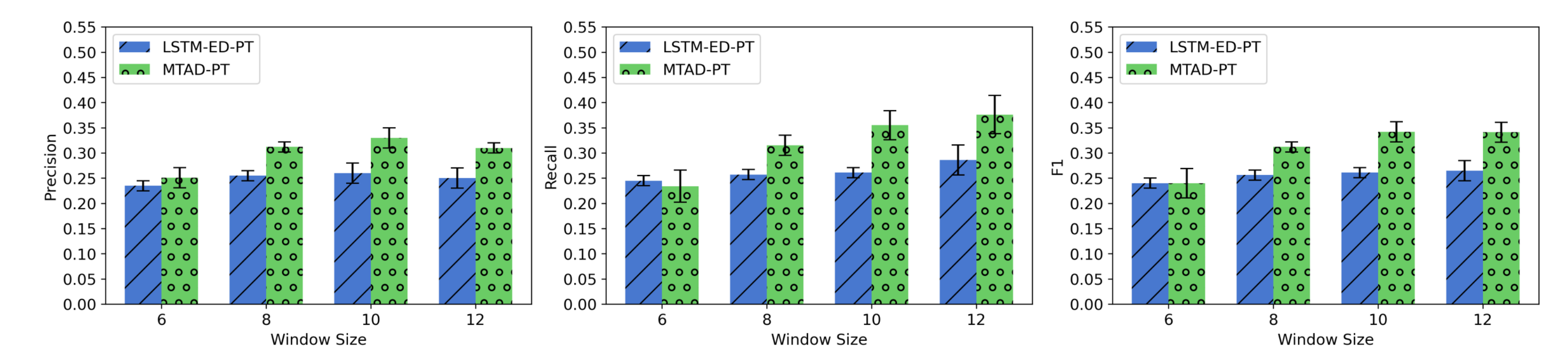}
  \caption{Comparison between LSTM-ED-PT and MTAD-PT at different window sizes.}
  \label{personal}
\end{figure*}
\subsubsection{Performance}\label{performance}
We evaluate the performance of our algorithm by comparing it with five state-of-the-art baselines for anomaly detection, namely, OCSVM, IF, LSTM-VAE, DAGMM, and LSTM-ED. As shown in Table \ref{tab:general}, MTAD performs significantly better than all traditional machine learning and deep learning methods in terms of P, R, and F1. Particularly, MTAD's 0.29 F1 score is 2.6 times greater than a standard LSTM autoencoder (LSTM-ED). Unlike other methods, DAGMM does not compute a normal event decision boundary, i.e., train only with normal event data. Consequently, we observe that DAGMM has a higher recall than methods like LSTM-ED, LSTM-VAE, and OCSVM, but it has poor precision. 

Interestingly, IF performs better than deep models like LSTM-ED and LSTM-VAE. IF directly predicts a rare event without considering temporal information, whereas LSTM approaches used windows that might contain unknown rare events or behavioral discrepancies, thus, resulting in poor performance. Moreover,  we observe that unsupervised LSTM autoencoder approaches are sensitive to variance in human behavior (Table \ref{tab:general}), suggesting that the latent representation computed might be biased to a specific user ``persona''. To address this, we attempt to personalize our approach to improve performance.  

\begin{table}[]
    \caption{Comparison of MTAD with baselines using precision (P), Recall (R), and F1-score (F1).}
    \centering
    \begin{adjustbox}{max width=\linewidth}
    \begin{tabular}{llll}
    \toprule
     Algorithm & P (std) & R (std) & F1 (std)  \\
     \midrule
     OCSVM & 0.32 (0.04) &0.07 (0.00) &0.12 (0.06)\\
     IF &0.22 (0.01) & 0.18 (0.01) & 0.20 (0.00)\\
     LSTM-VAE &0.28 (0.04) &0.07 (0.01) &0.12 (0.01)\\
     DAGMM &0.04 (0.01) &0.11 (0.02) &0.06 (0.01)\\
     LSTM-ED &0.25 (0.03) &0.07 (0.01) &0.11 (0.01)\\
     MTAD &\textbf{0.47 (0.05)} &\textbf{0.21 (0.03)} &\textbf{0.29 (0.04)}\\
    \bottomrule
    \end{tabular}
    \end{adjustbox}
    \label{tab:general}
\end{table}

\subsubsection{Personalization}\label{personalresults}

 Towards personalization, we applied within-subject normalization to capture user-specific behavior changes and computed user-specific thresholds $\gamma^u$ from the individual's validation data. Detecting rare events using these personalized thresholds (PT) yields performance improvements for all threshold-based methods, as shown in Table \ref{tab:personal}. Overall, MTAD-PT is the best performing method with an F1 of 0.34, a 0.05 increase from the general model. From Table \ref{tab:general} and \ref{tab:personal}, we see that unsupervised methods like LSTM-VAE-PT and LSTM-ED-PT show the most F1 score improvements of 0.14 and 0.15, respectively. Interestingly, by personalizing MTAD, we observe a trade-off between precision and recall. Additionally, methods like IF directly predict an anomaly without thresholds and cannot be personalized without training. Our experiments show that methods like MTAD can achieve performance improvement simply by personalized thresholding without additional training, demonstrating its advantage in human-centered problems. 
\begin{table}[]
    \caption{Comparison of personalized threshold models using precision (P), Recall (R), and F1-score (F1).}
    \centering
    \begin{adjustbox}{max width=\linewidth}
    \begin{tabular}{llll}
    \toprule
     Algorithm & P (std) & R (std) & F1 (std)  \\
     \midrule
     LSTM-VAE-PT &0.25 (0.02) &0.27 (0.02) &0.26 (0.02) \\
     LSTM-ED-PT &0.26 (0.02) &0.26 (0.02) &0.26 (0.01)\\
     MTAD-PT &\textbf{0.33 (0.02)} &\textbf{0.35 (0.03)} &\textbf{0.34 (0.03)}\\
    \bottomrule
    \end{tabular}
    \end{adjustbox}
    \label{tab:personal}
\end{table}
\vspace{-0.4cm}

\subsubsection{Effect of parameters}\label{effectofparameters}
\noindent\textbf{Window size.} Optimal window size is critical to sufficiently discerning differences between normal and rare events. Smaller window sizes allow each day to have more impact, whereas larger ones spread importance across many days. Thus, we evaluate the performance of MTAD, LSTM-ED, MTAD-PT, LSTM-ED-PT using four different window sizes $l \in \{6, 8, 10, 12\}$. 


We observe several interesting trends from Figures \ref{general} and \ref{personal}. First, the performance of MTAD and MTAD-PT increases with larger window sizes from 6 to 10. However, the difference in F1 score of MTAD-PT at $l=10$ and $l=12$ (0.34 vs 0.34) is insignificant. Second, the performance of LSTM-ED deteriorates gradually with increasing window sizes. Conversely, LSTM-ED-PT F1 score increases, demonstrating the robustness of using user-specific thresholds. Third, by personalizing, we observe a trade-off between P and R for MTAD at all window sizes. For our problem, the higher recall is acceptable, and we use a window size $l=10$. \\
\noindent\textbf{Decay constant.} The constant $\lambda$ used in exponential weighting determines the intensity of decay where higher values drastically reduce the weight of days farther from the current day as opposed to lower values. We evaluated the sensitivity of MTAD and MTAD-PT at different decay constants $\lambda \in \{0.5, 2, 5, 10\}$ and identified no significant changes (Appendix \ref{apd:3}). Intuitively, we expect this behavior as the proposed method only magnifies the anomaly score of windows with at least one rare event, leaving windows with normal events unchanged. 

\subsubsection{Utility of the sequence predictor}\label{ablation}
 We evaluate the necessity of having both tasks by performing an ablation study at inference. After training, we treat the sequence predictor as a supervised classifier where a $W$ has a rare event at time $t$ if the transition vector value $r_{t-1} = 1$, i.e., there is a workplace performance change between the two most recent days. The model obtained a P, R, and F1 of 0.52, 0.04, and 0.07 respectively. An 0.07 F1 score of the sequence predictor is poor compared to the baselines in Table \ref{tab:general}. Additionally, this experiment shows that a combined multi-task model has superior performance compared to standalone methods.


\subsubsection{Analyzing event type and valence} \label{eventtype}
We analyze the events identified by our model and observe that it is capable of identifying personal, work, health, and financial events (Table \ref{tab:events}). These types of events directly affect the participant or their relatives. Our method is unable to identify societal and miscellaneous events. These events could be related to politics, sports, or other local activities which could indirectly affect mood. From Table \ref{tab:events}, we observe that MTAD-PT is fairly balanced at detecting positive and negative events with similar recall (0.40 vs 0.39).
\begin{table}[h]
    \caption{Events detected or recall (R) using MTAD-PT distributed by type}
    \centering
    \begin{adjustbox}{max width=\linewidth}
    \begin{tabular}{lll}
    \toprule
     Event Type & Total Events & Events Detected (R)  \\
    \midrule
    \multicolumn{3}{l}{\cellcolor[HTML]{EFEFEF}\textit{Type}} \\                                      
     Personal & 92 & 41 (0.44)  \\
     Work & 69 & 21  (0.30)\\
     Health & 14 & 7 (0.50) \\
     Financial & 13 & 9 (0.69) \\
     Societal & 8 & 0 (0.00) \\
     Other & 2 & 0  (0.00)\\
    \multicolumn{3}{l}{\cellcolor[HTML]{EFEFEF}\textit{Valence}} \\        
    Positive &136 &54 (0.40)\\
    Negative &62 &24 (0.39)\\
    \bottomrule
    \end{tabular}
    \end{adjustbox}
    \label{tab:events}
\end{table}
\section{Discussion} \label{sec:disc}
\subsection{Summary of results}
Initially, granger causality testing suggested that location features are crucial in detecting behavioral changes after an event. Perhaps, both negative and positive events such as loss of loved one or vacation result in changes in location dynamics. We observed that 46 time series were significant for location distance between 12am-6am. This is considerably large from an extreme anomaly detection perspective. Thus, motivating us to build a multi-task learning model to detect rare LEs.

The results of rare life event detection presented in section \ref{sec:results} highlight the advantages our deep learning approach. A multi-task setup can be used to overcome the deficiencies of a purely unsupervised deep learning model. Specifically, the presence of a severe class imbalance ($<2\%$) can be addressed using our method. In comparison to \citet{burghardt2021having}, we achieve a comparable F1 of 0.34 on a more challenging problem because of the aforementioned class imbalance (11.7\% \& 14.9\% vs. 1.9\%). Moreover, our method can detect both positive and negative LEs in addition to different events types such as personal, work, health, and financial \citep{burghardt2021having}. Our approach can be extended to other applications by appropriately identifying auxiliary tasks to positively transfer knowledge to the main task.

With the future vision of deploying models on mobile devices, it is imperative that models are not re-trained on the phone. Higher computational costs of training models results in slower application performance. Consequently, human interaction with the application might be reduced. Personalizing the thresholds for each individual without re-training addresses this issue while simultaneously improving the performance of MTAD and other unsupervised models.  

\subsection{Implications} \label{sec:implications}
 Two major directions could greatly benefit from our work. First, as LEs are difficult to identify without explicit questions, detecting them using mobile phones is valuable. Second, interventions can alleviate the stressful impact of LEs.\\
\noindent\textbf{Detection.} Generally, LEs are identified only through self-reports. Automated detection in-the-wild is difficult because of the subtleties of human behavior. Our analysis of data from Android and iOS devices illustrates the effectiveness of using passive sensing data instead of conducting monthly interviews or similar alternative methods with employees. Thus, mobile phones are a low-burden option offering continuous real-time sensing to detect rare LEs. Ultimately, we envision detection leading to helpful and empathetic interventions.\\ 
\noindent\textbf{Ubiquitous health interventions.}
Individuals can use our detection algorithm for teletherapy and self-monitoring. In teletherapy, smartphone sensing applications can connect people to counselors, mentors, or therapists offering event-specific services. For example, a death in the family requires the expertise of a grief counselor, whereas a mentor can help tackle the stressors of a new job promotion. Applications like Talkspace and Betterhelp offer several services in this sector, and our methods can be integrated with similar applications. Second, our algorithm can be extended for self-monitoring, where an app tracks anomalous behaviors longitudinally—ultimately suggesting intervention strategies for significant behavioral deviations.

Organizations should be proactive in improving the mental health and wellness of its employees. Here, we describe three intervention scenarios using smartphones. First, having helpful co-workers reduces negative triggers. Our method may be used to detect a life event passively and provide incentives to an information worker to help colleagues. Second, \citet{spector2002emotion} suggest that emotion about control over a stressful event affects health. Thus, emotion regulation strategies such as cognitive re-appraisal (re-interpreting a situation) can be prompted to the user through mobile apps \citep{katana2019emotion}. Third, organized leisure crafting have been shown to positively affect mental health and can be used as an intervention tool \citep{ugwu2017contribution}.

\subsection{Limitations}
Some limitations of this work should be noted. First, we do not address the ``cold-start problem'' to evaluate performance for an unseen user. Thus, our model with personalization requires user behavioral data to construct specific thresholds and latent spaces for detecting LEs in the future. Second, it is useful to understand how the various mobile features contribute the most to detection. The latent features constructed by autoencoders are unobserved feature spaces compressed into a low-dimensional for reconstruction and prediction. Therefore, interpretation of these features is not straight-forward, the additional scaling of the auxiliary task further hinders this ability. Third, some rare events cannot be detected because of their similarity to normal events. In essence, there are several confounding factors that may or may not elicit a behavioral change in an individual. For example, if a person's responsibilities are similar after a job promotion, their routine and actions might not be significantly different. Conversely, it is also possible that normal days are anomalies and not related to the events themselves. 

\subsection{Ethical considerations} \label{sec:ethics}
 While monitoring worker behavior has benefits to health, it also highlights the need for ethical data usage. For instance, organizations analyzing mobile phone data should use informed consent to protect private data. The primary intention of life event detection must be to offer help and support. Nevertheless, sensing data can be used adversarially to monitor employee productivity to maximize benefits for the organization. 
 Thus, sacrificing trust between employee and organization, while damaging interaction between people and technology. We do not advocate the use of mobile sensing methods like event detection in these scenarios. Future studies must collect data only through transparent and ethical processes to protect the employee's data. Moreover, extreme anomaly detection have higher error rates owing to its challenging nature. Therefore, having a human-in-the-loop is a necessity. We discuss two scenarios of good and bad interventions.\\ 
 \noindent\textbf{Scenario 1.} A recently promoted information worker is overwhelmed and unable to meet product goals. A \textbf{good} intervention offers resources for mentorship and stress management. A \textbf{bad} intervention gives ultimatums that affect job security. \\
 \noindent\textbf{Scenario 2.} An employee recovering from a physical injury struggles to keep up with their old workload. A \textbf{good} intervention connects them to an accessbility expert or counselor to help them with their specific issues. A \textbf{bad} intervention monitors the employees performance and puts them on performance review.

\section{Conclusion}
In this paper, we showed that mobile sensing can address the challenging task of detecting rare LEs in-the-wild. We proposed MTAD, a multi-task framework for life event detection using behavioral sensing data from mobile phones. MTAD's use of an auxiliary sequence predictor addresses several challenges like extreme class imbalance ($<2\%$) and biased reconstruction score. We demonstrated the superior performance of our approach using a real-world longitudinal dataset by comparing it with state-of-the-art baselines. From a human-centered perspective, MTAD's effectiveness in personalization without additional training, robustness to decay, and balanced prediction of positive and negative events are desirable qualities. Ultimately, we envision our work motivates ubiquitous health-based intervention strategies through smartphones.


\acks
This work is supported in part by the Army Research Laboratory (ARL) under Award W911NF202011. The views and conclusions contained herein are those of the authors and should not be interpreted as necessarily representing the official policies, either expressed or implied by ARO or the U.S. Government.  
\bibliography{jmlr-sample}

\appendix
\section{}\label{apd:first}
\subsection{Implementation details}
An open source version of the Tesserae Dataset is available on request \footnote{https://tesserae.nd.edu/}, additional data will be released in the future.
All experiments were performed on an NVIDIA RTX 3070 8 GB laptop GPU. One-Class Support Vector Machine (OCSVM) and Isolation Forest (IF) were implemented using scikit-learn\footnote{https://scikit-learn.org/stable/modules/\newline generated/sklearn.svm.OneClassSVM}, \footnote{
https://scikit-learn.org/stable/modules/\newline generated/sklearn.ensemble.IsolationForest}. We implemented the deep learning models using keras and tensorflow. LSTM-ED, LSTM-VAE\footnote{https://github.com/twairball/keras\_lstm\_vae}, and DAGMM\footnote{
https://github.com/tnakae/DAGMM}  were designed using open source repositories and articles\footnote{https://blog.keras.io/building-autoencoders-in-keras.html}. 
We use the following packages in our MTAD implementation:
\begin{itemize}
    \item python == 3.9.7
    \item numpy == 1.21.3
    \item scikit-learn == 1.0
    \item pandas == 1.3.4
    \item keras == 2.4.3
    \item tensorflow == 2.5.0
\end{itemize}

\subsection{MTAD hyperparameters}
MTAD is trained using the following hyperparameters: epochs = 500; batch size (bs) = 128; Adam optimizer with a learning rate = $1 \times 10^{-4}$. In our implementation, we use window size = 10 for the 29 features. We set two early-stopping criteria to prevent overfitting: (1) if validation loss has not improved in 10 epochs, and (2) if validation loss reaches 0.2. The reason for criteria (2) is to account for the reduced data set size when window size is increased. Additionally, the class weights for the cross-entropy loss were calculated using the  compute\_sample\_weight function from sklearn.utils.class\_weight.

\begin{table}[!ht]
\caption{MTAD architecture and parameters}
\begin{adjustbox}{max width = \linewidth}
\begin{tabular}{@{}lllll@{}}
\textbf{Layer} & \textbf{Parameters}   & \textbf{Output shape}\\ \bottomrule
\multicolumn{3}{l}{\cellcolor[HTML]{EFEFEF}\textit{Encoder}} \\         
\midrule
    \multirow{3}{*}{LSTM} &units=100 &\\
    &activation=relu &(bs, 100)\\
    &recurrent dropout=0.2 &\\
    Repeat Vector &n=10 &(bs, 10, 100)  \\
\midrule
\multicolumn{3}{l}{\cellcolor[HTML]{EFEFEF}\textit{Decoder}}\\
\midrule
    \multirow{2}{*}{LSTM} &units=100 &(bs, 10, 100)\\
    &activation=relu &\\
    Dense &units=29 &(bs, 10, 29)  \\
\midrule
\multicolumn{3}{l}{\cellcolor[HTML]{EFEFEF}\textit{Predictor}}\\
\midrule
    \multirow{3}{*}{LSTM} &units=100 &\\
    &activation=relu &(bs, 10, 100)\\
    &recurrent dropout=0.2 &\\
    \multirow{2}{*}{Dense} &units=8 &(bs, 10, 8)  \\
    &activation=softmax & \\
\bottomrule
\end{tabular}
\end{adjustbox}
  \label{tab:architecture}
\end{table}

\subsubsection{Hyper-parameters of other models}
Here, we list the best parameters obtained from tuning that maximized performance. 

\noindent\textbf{One Class SVM.} kernel='rbf'; gamma=0.01; nu=0.03.

\noindent\textbf{Isolation Forest.} n\_estimators=200 and contamination=0.02 

\noindent\textbf{LSTM-ED.} We use the same encoder and decoder architecture and parameters (Table \ref{tab:architecture}) as Task A of MTAD and train the model for 300 epochs. 

\noindent\textbf{LSTM-VAE.} Here, our encoder uses an LSTM layer with units=100 and activation=relu. Next, the latent representation is computed through a distribution by sampling. We use two Dense layers with 32 units for this. And the Repeat Vector function copies the latent vector 10 times based on the window size. The decoder uses the same architecture and parameters as MTAD's decoder (Table \ref{tab:architecture}).

\noindent\textbf{DAGMM.} Compression Network Units: $100 \rightarrow 50 \rightarrow 20 \rightarrow 10$; activation=tanh \\
Estimation Network: $5 \rightarrow 10 \rightarrow 2$; activation=tanh; \\dropout=0.5 \\
We train using the following setup: batch size = 256, epochs = 9000; learning rate= 0.0001; lambda1=0.01;lambda2=0.00001

\subsection{Demographics}\label{apd:second}
\begin{table}[h]
\caption{Demographic descriptors of participants in our analysis}
\begin{adjustbox}{max width = \linewidth}
\begin{tabular}{p{0.6\linewidth}ll}
\textbf{Category} & \textbf{Count}   & \textbf{Percentage}\\ \bottomrule
\multicolumn{3}{l}{\cellcolor[HTML]{EFEFEF}\textit{Sex}} \\                                      
Female & 59 &46.8\%   \\
Male   & 67 &53.2\%    \\
\multicolumn{3}{l}{\cellcolor[HTML]{EFEFEF}\textit{Occupation}}\\
Architecture and Engineering &16 &12.7\% \\
Business and Financial Operation &28 &22.2\% \\
Computer and Mathematical &43 &34.1\% \\
Construction and Extraction &1 &0.8\% \\
Education, Training, and Library Services &1 &0.8\% \\
Healthcare Practitioners and Technical Healthcare Occupations &2 &1.6\% \\
Healthcare Support &1 &0.8\% \\
Management &13 &10.3\% \\
Office and Administrative Support Occupations &3 &2.3\% \\
Production &1 &0.8\% \\
Sales and Related Occupations &3 &2.3\% \\
Other &14 &11.1\% \\
\multicolumn{3}{l}{\cellcolor[HTML]{EFEFEF}\textit{Education}}\\
High school degree &1 &0.8\% \\
College degree &58 &46.0\% \\
Master's degree &38 &30.1\% \\
Doctoral degree &2 &1.6\% \\
Some high school &1 &0.8\% \\
Some college &14 &11.1\% \\
Some graduate school &12 &9.5\% \\
\bottomrule
\end{tabular}
\end{adjustbox}
  \label{tab:demographics}
\end{table}
\newpage
\subsection{Decay Constant} \label{apd:3}
\begin{figure}[h]
    \centering
    \includegraphics[width=\linewidth]{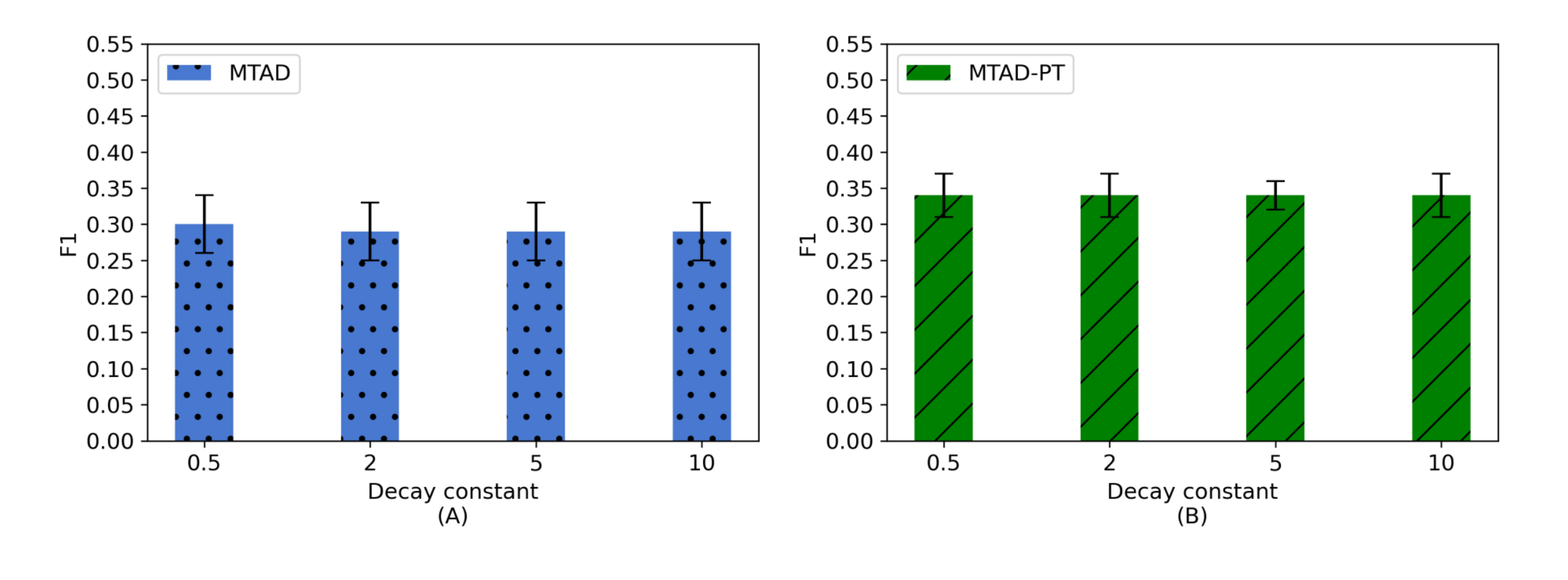}
    \caption{Comparing different decay constant values: (A) F1 score for MTAD (B) F1 score for MTAD-PT.}
    \label{decay}
\end{figure}

\end{document}